\documentclass[letterpaper]{article} % DO NOT CHANGE THIS
\usepackage[preprint]{aaai2027}  % DO NOT CHANGE THIS
\usepackage[hyphens]{url}  % DO NOT CHANGE THIS
\usepackage{graphicx} % DO NOT CHANGE THIS
\usepackage{natbib}  % DO NOT CHANGE THIS AND DO NOT ADD ANY OPTIONS TO IT
\usepackage{caption} % DO NOT CHANGE THIS AND DO NOT ADD ANY OPTIONS TO IT
\usepackage{algorithm}
\usepackage{algorithmic}
\usepackage{multirow}
\usepackage{enumitem}
\usepackage[figuresright]{rotating}
\usepackage[table]{xcolor}
\usepackage{booktabs}
\usepackage{array}
\usepackage{makecell}
\usepackage{multirow}
\usepackage{pdflscape}
\usepackage{amssymb}
\usepackage{amsmath}
\usepackage{newfloat}
\usepackage{listings}
\DeclareCaptionStyle{ruled}{labelfont=normalfont,labelsep=colon,strut=off} % DO NOT CHANGE THIS
\floatstyle{ruled}
\newfloat{listing}{tb}{lst}{}
\floatname{listing}{Listing}

\usepackage{booktabs}

\title{AtomBridge: Agentic VLA Inference Plugin for Long-Horizon Tasks in Scientific Experiments}

\author{
    Yiwen Pang\textsuperscript{\rm 1,2}, \ Bo Zhou\textsuperscript{\rm 1,2}, \ Changjin Li\textsuperscript{\rm 1,2}, \ Xuanhao Wang\textsuperscript{\rm 1,2}, \ Shengxiang Xu\textsuperscript{\rm 1,2}, Deng-Bao Wang\textsuperscript{\rm 1,2}, \ Peng Cheng\textsuperscript{\rm 3}, \ Shimin Di\textsuperscript{\rm 1,2}, \ Jingkuan Song\textsuperscript{\rm 3}, \ Min-Ling Zhang\textsuperscript{\rm 1,2}
}
\affiliations{
    \textsuperscript{\rm 1}School of Computer Science and Engineering \& School of Software Engineering \\
    \& School of Artificial Intelligence, Southeast University, Nanjing, China\\
    \textsuperscript{\rm 2}Pattern Learning and Mining Lab, Southeast University, Nanjing, China\\
    \textsuperscript{\rm 3}Tongji University, Shanghai, China\\
    shimin.di@seu.edu.cn
}

\begin{document}

\maketitle

\begin{abstract}
Robotic laboratories play a critical role in autonomous scientific discovery by enabling scalable, continuous experimental execution. Recent vision–language–action (VLA) models offer a promising foundation for robotic laboratories. However, scientific experiments typically involve long-horizon tasks composed of multiple atomic tasks. Existing VLA models may fail to perform composed tasks formed by reordering and composing these known atomic actions. This limitation can arise from a skill-chaining gap caused by robot-state mismatch: the terminal robot state of one skill can fall outside the valid initial-state distribution of the next. To address this challenge, we propose AtomBridge, an Agentic VLA Inference Plugin for Long-Horizon Tasks in Scientific Experiments. AtomBridge attaches at inference time to a VLA policy already fine-tuned on atomic tasks, while keeping its weights fixed.  At each task boundary, it uses LLM-based transition reasoning and robotic-action code generation to insert transitional actions between consecutive tasks. This plug-and-play design mitigates the skill-chaining gap caused by robot-state mismatch without additional VLA fine-tuning or demonstrations of composed long-horizon sequences. Across scientific manipulation sequences in simulation and a real-world experimental environment, AtomBridge improves execution continuity and per-step atomic-task success. On 8-step composed tasks, AtomBridge improves full-sequence success by 10\%\textasciitilde25\%.
\end{abstract}

\begin{figure*}[!t]
    \centering
    \includegraphics[width=\linewidth]{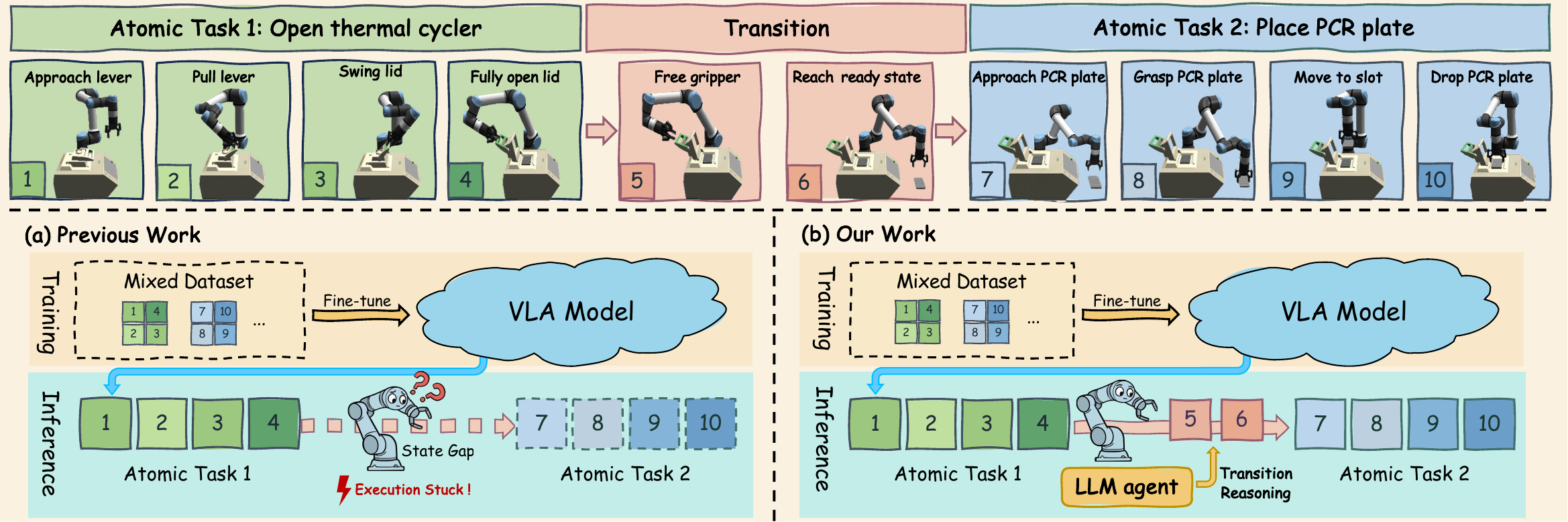}
    \caption{
    Given a VLA model fine-tuned for atomic tasks in scientific experiments,
    it may fail to complete the composed task because of a robot-state mismatch between consecutive atomic tasks.
    In this paper,
    we propose an LLM-based agentic inference plugin to generate transitional robotic execution code, improving execution of the long-horizon tasks.
    }
    \label{fig:motivation}
\end{figure*}

\section{Introduction}
\label{sec:intro}
In recent years, the AI4Science paradigm~\cite{wang2023scientific} has evolved from single-model approaches~\cite{jumper2021highly,bi2023accurate} toward autonomous scientific discovery~\cite{novikov2025alphaevolve,volk2023alphaflow,dai2025adaptive,koscher2023autonomous}.
Emerging research aims to directly embed artificial intelligence into scientific workflows to systematically enhance the entire research lifecycle~\cite{koscher2023autonomous,szymanski2023autonomous,boiko2023autonomous}. 
A robotic laboratory plays a critical role in enabling autonomous discovery~\cite{burger2020mobile,szymanski2023autonomous,coley2019robotic}.
It creates experimental environments with high-precision control, strong reproducibility, and large-scale execution capabilities, substantially accelerating scientific progress while reducing labor and time costs.
Moreover, robots capable of continuous, unattended operation significantly reduce researchers’ exposure to hazardous experimental conditions.
Autonomous laboratories have successfully discovered correct synthesis routes for 41 theoretically predicted materials through large-scale automated material synthesis experiments~\cite{szymanski2023autonomous}.

Current robotic laboratories typically follow the automation paradigm based on manually programmed, fixed experimental procedures
~\cite{coley2020autonomous}.
This choice is largely motivated by the high-precision requirements of scientific experiments.
For instance, in high-throughput chemical screening~\cite{macarron2011impact},
a single experimental cycle typically consists of sample preparation and reaction execution, for which fixed programs for robotic arms provide stability guarantees.
However, such rigid workflows limit the robot’s ability to reconfigure task sequences or respond to unforeseen situations, making them ineffective in open-ended experimental settings~\cite{ochiai2025automating}.
As an example, the parameters of a thermal cycler—such as cycle numbers, temperatures, and durations—must be flexibly adjusted according to experimental objectives, templates, and primers (the scientific instrument in Fig.~\ref{fig:motivation}), and thus cannot be pre-encoded as a single robot trajectory.

Recent vision–language–action (VLA) models~\cite{black2024pi_0,kim2025openvla} provide a flexible interface for reconfigurable laboratory manipulation.
Unlike fixed-program systems, this paradigm allows robots to interpret natural-language instructions and switch among learned manipulation skills.
As the set of atomic experimental procedures is generally fixed and finite, VLA models can be trained or fine-tuned using data collected only for \textit{atomic tasks}.
Then, to support diverse experimental protocols, these atomic tasks can be dynamically composed and reordered.  
However, we observe a phenomenon with the VLA model when performing these long-horizon tasks composed of atomic tasks:
\textit{
even though the VLA model fine-tuned on atomic tasks can handle these atomic tasks smoothly, it still cannot correctly complete composed tasks made up of these known atomic tasks.}
As illustrated in Fig.~\ref{fig:motivation} (a), a fine-tuned VLA model can independently execute \texttt{open thermal cycler} and \texttt{place PCR plate}, but fails to correctly execute the composed task \texttt{open thermal cycler–place PCR plate}.
This limitation can reflect a \textit{skill-chaining gap} caused by robot-state mismatch~\cite{konidaris2009skill,lee2022adversarial,jung2025planning}: although the VLA model has learned each atomic task, the terminal robot state of the pre-task can fall outside the valid initial-state distribution of the post-task, preventing the required inter-skill transition.

To address this challenge, we propose AtomBridge, an agentic, plug-and-play inference plugin for long-horizon scientific manipulation.
This bridger primarily inserts transitional actions between consecutive atomic tasks within a long-horizon task sequence, thereby ensuring smoother task transitions.
Our contributions are as follows:
\begin{itemize}
    \item We identify the skill-chaining gap caused by the absence of explicit transitions between atomic tasks, and introduce the \emph{ready state}: a pre-manipulation state that lies within the valid initial-state distribution of the next atomic task.
    \item We introduce AtomBridge, an agentic and plug-and-play inference plugin that detects task completion, retrieves the next task's ready state, and generates constrained transition actions. It attaches to a fixed atomic-task VLA policy without collecting additional data or re-training the VLA.
    \item In our simulation test, AtomBridge improves 53 of 60 post-initial atomic-task comparisons and obtains full-sequence success in 11 of 12 task-model configurations, whereas all matched baselines obtain zero. In our real thermal-cycler experiment, it increases the five-step full-sequence success rate from 0\% to 40\%.
\end{itemize}

\begin{figure*}[!t]
  \centering
  \includegraphics[width=\linewidth]{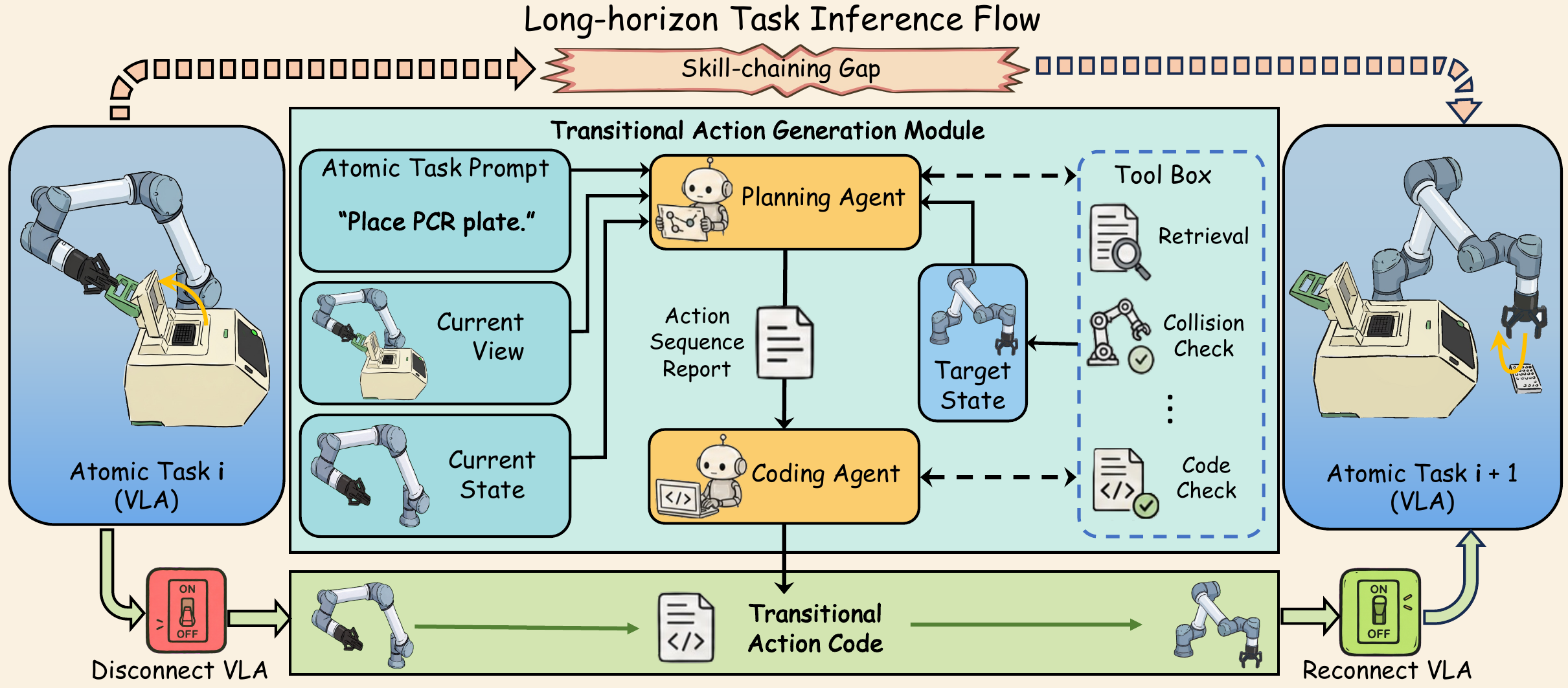}
  \caption{Overview of AtomBridge.  Fixed VLA policies execute the atomic tasks in a long-horizon sequence.  At each task boundary, the intervention switch detects completion or a task-specific timeout. AtomBridge then retrieves a pre-contact ready state for the next task from its atomic-task memory, and agents will plan and generate transitional action code.  After executing the transition, the next atomic VLA policy resumes.  This completion--transition--resume cycle repeats without modifying the VLA policy.}
  \label{figs:pipeline}
\end{figure*}

\section{Related Work}
\label{sec:related}
\noindent
\textbf{Robotic Labs.}
Robotic Labs combine manipulators, mobile platforms, and automated instruments to reduce manual experimental labor~\cite{boyd2002robotic}.
Representative systems automate thin-film research and high-throughput perovskite preparation~\cite{macleod2020self,zhao2021discovery}, use language models and active sensing to plan solid-state synthesis~\cite{szymanski2023autonomous}, and explore photocatalytic parameter spaces with mobile robots~\cite{burger2020mobile}.
Earlier systems such as Adam further coupled automated experiments with hypothesis testing to identify gene functions~\cite{king2009automation}.

\noindent
\textbf{Vision-Language-Action Models.}
VLA models map visual observations and language instructions to robot actions, using vision-language representations~\cite{zhang2024vision,radford2021learning} to support generalization across instructions and scenes.
Diffusion Policy~\cite{chi2025diffusion} and RDT~\cite{liurdt}
generate continuous action trajectories through iterative denoising, while $\pi_0$~\cite{black2024pi_0} and $\pi_{0.5}$~\cite{black2025pi_} use flow matching with a PaliGemma backbone~\cite{beyer2024paligemma}.
In particular, LabVLA~\cite{ren2026labvla} is a VLA model trained specifically for scientific laboratory scenarios.
After large-scale robot pretraining and task-specific fine-tuning, these models can execute diverse manipulation skills, but composing independently trained skills remains difficult.

\noindent
\textbf{Long-Horizon Challenges.}
Long-horizon manipulation faces multiple challenges.  
First, small perception or control errors can accumulate across many time steps, so later actions begin from increasingly inaccurate states~\cite{10400863}. 
Closed-loop feedback~\cite{pmlr-v87-ebert18a}, replanning~\cite{10328058}, and recovery~\cite{pmlr-v205-sung23a} are used to correct these deviations during execution.  
Second, a high-level goal must be decomposed into executable atomic tasks.  
Hierarchical policies make this decomposition explicit~\cite{dietterich2000hierarchical}, while recent VLAs learn task decompositions, subtask tokens, or phase-aware representations: $\pi_{0.5}$ adds task-decomposition pairs during vision--language pretraining~\cite{black2025pi_}. 
LohoVLA predicts atomic-task tokens before action tokens~\cite{yang2025lohovla}. 
LongVLA attends to different visual information across movement phases~\cite{fan2025long}.
Third, the resulting atomic policies must be connected reliably.  
This \textit{skill-chaining gap} arises when one policy's terminal-state distribution does not overlap with the next policy's valid initial-state distribution, producing out-of-distribution or physically invalid starting states.  
Existing solutions learn backward-reachable options~\cite{konidaris2009skill}, relay or goal-conditioned policies~\cite{gupta2020relay}, or state-distribution alignment~\cite{clegg2018learning,lee2022adversarial}. 
Alternatively, they progressively fine-tune sub-policies~\cite{chen2023sequential} or plan over learned skill priors and connector policies at inference time~\cite{mishra2023generative,jung2025planning}.  
These approaches address decomposition or state compatibility, but do not explicitly generate task-conditioned transitions between fixed VLA policies.

\begin{figure*}[!t]
  \centering
  \includegraphics[width=\linewidth]{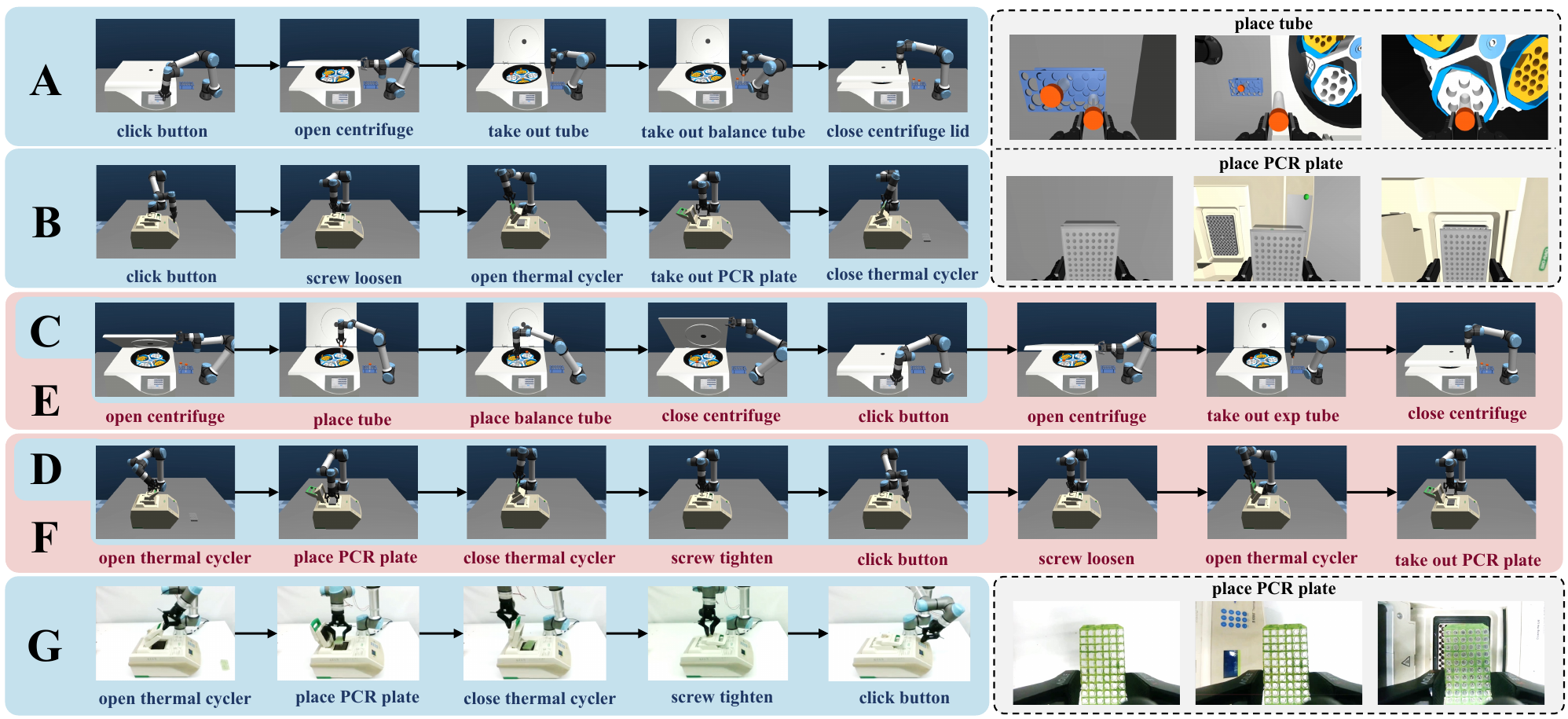}
  \caption{
  Definitions of the long-horizon scientific operation tasks. Tasks A, B, C, D and G are long-horizon task sequences of length 5, whereas Tasks E and F are long-horizon task sequences of length 8. The atomic tasks inside the dotted box require higher operation precision.}
  \label{figs:scitask}
\end{figure*}

\section{AtomBridge: Agentic Plugin}
\label{sec:method}
Let $\mathcal{A}=\{\tau^j,\ldots,\tau^k\}$ be the set of atomic task descriptions, and let $\pi$ be a fixed VLA policy trained only on demonstrations of tasks in $\mathcal{A}$.  
A long-horizon task of length $N$ is an arbitrary ordered sequence $\mathcal{T}=[\tau_1,\ldots,\tau_N]$ with $\tau_i\in\mathcal{A}$. 
For a demonstration trajectory of task $\tau$, let $\mathcal{D}_{\tau}=\{(I_t,s_t)\}_{t=0}^{L-1}$, where $I_t$ is the RGB image frame at time step $t$, $s_t$ is the corresponding robot state, and $L$ is the trajectory length. 
Let $r$ be the last frame before the robot begins the task-specific operation.  
Its \emph{ready-state set} is
\begin{equation}
    \mathcal{R}_{\mathcal{D}}(\tau)
    =\{s_t\mid 0\leq t\leq r\}.
\end{equation}
Thus, a ready state is any state in which the task has not yet begun and the robot has not started to manipulate the target.  
Ready state is inherently non-unique and visually gradual. 
Pooling these sets across demonstrations induces the empirical ready-state distribution $p_{\mathcal{D}}^{\mathrm{ready}}(s\mid\tau)$. 
When $\pi$ executes $\tau_i$, its terminal state $s_{\mathrm{cur}}$ is immediately passed to the next policy invocation:
\begin{equation}
    \begin{aligned}
    s_{\mathrm{cur}} &\sim p_{\pi}^{\mathrm{end}}(s\mid\tau_i)
    \quad\longrightarrow\quad
    \pi(\tau_{i+1};s_{\mathrm{cur}}),\\
    p_{\pi}^{\mathrm{end}}(s\mid\tau_i) &\neq p_{\mathcal{D}}^{\mathrm{ready}}(s\mid\tau_{i+1}).
    \end{aligned}
\end{equation}
where $p_{\pi}^{\mathrm{end}}$ is the terminal-state distribution induced by $\pi$. 

The mismatch between $p_{\pi}^{\mathrm{end}}(s\mid\tau_i)$ and $p_{\mathcal{D}}^{\mathrm{ready}}(s\mid\tau_{i+1})$ is the skill-chaining gap. 
We define the required \emph{transition} as an action sequence $\mathbf{a}^{\mathrm{tr}}_i$ that moves the robot to a valid pre-contact state for $\tau_{i+1}$:
\begin{equation}
    s_{\mathrm{cur}} \xrightarrow{\;\mathbf{a}^{\mathrm{tr}}_i\;} s_{\mathrm{ready}},
    \qquad
    s_{\mathrm{ready}}\sim p_{\mathcal{D}}^{\mathrm{ready}}(s\mid\tau_{i+1}).
\end{equation}
AtomBridge retrieves $s_{\mathrm{ready}}$ from an independent demonstration of $\tau_{i+1}$ and generates $\mathbf{a}^{\mathrm{tr}}_i$ only at task boundaries. 
It therefore composes fixed VLA policies without additional VLA training (Fig.~\ref{figs:pipeline}).

\subsection{Intervention Switch}
At discrete observation step $t$, the switch receives the initial image frame $I_0$, current image frame $I_t$, and task description $\tau_i$. 
We use a frozen CLIP image-text encoder with trainable parameters $\theta$ outputs the completion probability $p_t=f_\theta(I_0,I_t,\tau_i)$. 
Let $w$ be the required number of consecutive positive predictions, $\eta$ the completion threshold, $H_i$ the maximum execution time for $\tau_i$.  The binary switch decision $z_t$ is activated when the latest $w$ predictions exceed $\eta$, or when $t\geq H_i$:
\begin{equation}
z_t =
\begin{cases}
1, & \begin{gathered}
     \left(\bigwedge_{j=t-w+1}^{t} p_j\geq\eta\right)
     \lor \left(t\geq H_i\right),
     \end{gathered}\\
0, & \text{otherwise}.
\end{cases}
\end{equation}
Here, $j$ indexes the prediction steps in the latest window.  The temporal condition suppresses transient false positives. 
When $z_t=1$, AtomBridge pauses the current policy, executes $\mathbf{a}^{\mathrm{tr}}_i$, resets the history, and starts the next policy. 
Training labels are supplied by task-specific physics success predicates, and splits are made by complete trajectory. 
Model and training details are deferred to the supplementary material.

\subsection{Transition Generation}
For each atomic task, AtomBridge stores one offline demonstration containing its description, observations, and robot states. 
This offline demonstration is called \textit{memory}, denoted by $\mathcal{M}$. 
For the next task, the ready-state retrieval tool searches its matched memory for a usable ready state close to the last pre-operation frame $r$:
\begin{equation}
    s_{\mathrm{ready}}=\operatorname{Retrieve}(\mathcal{M},\tau_{i+1}).
\end{equation}
A task-conditioned visual judge and a fast sliding-window search algorithm identify the onset of contact or manipulation. 
Given the next-task description and an earlier--later pair of memory frames, the task-conditioned visual judge determines whether either frame has exceeded ready state, for example through target contact, manipulation, or an unusably close gripper configuration.
The deterministic fast sliding-window search begins from a mid-trajectory window, accepts a non-exceeded later endpoint, or contracts and shifts the window left under a fixed judgment budget to return a usable pre-contact state near $r$.
The visual-judge prompt and the retrieval algorithm details are provided in the supplementary material.

Using the retrieved ready state together with calibrated camera views $V$ and calibration parameters $\kappa$, the planning agent determines the transition from $s_{\mathrm{cur}}$ to $s_{\mathrm{ready}}$.
The coding agent converts them into template Python commands:
\begin{equation}
    \mathbf{a}^{\mathrm{tr}}_i=\operatorname{Code}\!\left(\operatorname{Plan}(s_{\mathrm{cur}},s_{\mathrm{ready}},V,\kappa)\right).
\end{equation}
The host accepts only fixed motion commands (e.g., \texttt{open\_gripper}, \texttt{translate}, \texttt{rotate}) that satisfy schema, limit, transition-only, and collision checks. 
Implementation details are provided in the supplementary material.

\begin{table*}[!t]
    \setlength{\tabcolsep}{3mm}
    \centering
    \begingroup
    % \small
    \renewcommand{\arraystretch}{0.92}
    \begin{tabular}{c|l|cccccccc|c}
    \toprule
    \multicolumn{1}{c|}{\multirow{2}{*}{Task}}&\multicolumn{1}{l|}{\multirow{2}{*}{Method}}&\multicolumn{8}{c|}{Success Rate Per Atomic Task}&\multirow{2}{*}{\makecell{Full-Seq.\\Success}}\\
    \cline{3-10}
    &&1&2&3&4&5&6&7&8&\\
    \midrule

    \multirow{6}{*}{A}&$\pi_0$ (vanilla)&20/20&0/20&0/20&0/20&0/20&--&--&--&0/20\\
    &$\pi_{0.5}$ (vanilla)&20/20&0/20&0/20&0/20&0/20&--&--&--&0/20\\
    &$\pi_0$ (MP)&20/20&\underline{6/20}&0/20&0/20&0/20&--&--&--&0/20\\
    &$\pi_{0.5}$ (MP)&20/20&3/20&0/20&0/20&\underline{1/20}&--&--&--&0/20\\
    &$\pi_0$ (AtomBridge)&20/20&\textbf{20/20}&\underline{5/20}&0/20&\textbf{15/20}&--&--&--&0/20\\
    &$\pi_{0.5}$ (AtomBridge)&20/20&\textbf{20/20}&\textbf{10/20}&\textbf{5/20}&\textbf{15/20}&--&--&--&\textbf{3/20}\\
    \cline{1-11}
    \multirow{6}{*}{B}&$\pi_0$ (vanilla)&20/20&0/20&0/20&0/20&0/20&--&--&--&0/20\\
    &$\pi_{0.5}$ (vanilla)&20/20&0/20&0/20&0/20&0/20&--&--&--&0/20\\
    &$\pi_0$ (MP)&20/20&\underline{14/20}&3/20&0/20&1/20&--&--&--&0/20\\
    &$\pi_{0.5}$ (MP)&20/20&10/20&3/20&0/20&3/20&--&--&--&0/20\\
    &$\pi_0$ (AtomBridge)&20/20&\textbf{20/20}&\textbf{20/20}&\textbf{14/20}&\textbf{14/20}&--&--&--&\textbf{14/20}\\
    &$\pi_{0.5}$ (AtomBridge)&20/20&\textbf{20/20}&\underline{14/20}&\underline{10/20}&\underline{8/20}&--&--&--&\underline{7/20}\\
    \cline{1-11}
    \multirow{6}{*}{C}&$\pi_0$ (vanilla)&16/20&0/20&0/20&0/20&0/20&--&--&--&0/20\\
    &$\pi_{0.5}$ (vanilla)&18/20&0/20&0/20&0/20&0/20&--&--&--&0/20\\
    &$\pi_0$ (MP)&15/20&0/20&0/20&0/20&\underline{8/20}&--&--&--&0/20\\
    &$\pi_{0.5}$ (MP)&18/20&1/20&2/20&3/20&7/20&--&--&--&0/20\\
    &$\pi_0$ (AtomBridge)&16/20&\underline{6/20}&\underline{4/20}&\underline{14/20}&6/20&--&--&--&\underline{2/20}\\
    &$\pi_{0.5}$ (AtomBridge)&18/20&\textbf{13/20}&\textbf{12/20}&\textbf{18/20}&\textbf{12/20}&--&--&--&\textbf{5/20}\\
    \cline{1-11}
    \multirow{6}{*}{D}&$\pi_0$ (vanilla)&18/20&0/20&12/20&0/20&0/20&--&--&--&0/20\\
    &$\pi_{0.5}$ (vanilla)&18/20&0/20&12/20&0/20&0/20&--&--&--&0/20\\
    &$\pi_0$ (MP)&17/20&1/20&\underline{13/20}&\underline{13/20}&2/20&--&--&--&0/20\\
    &$\pi_{0.5}$ (MP)&20/20&4/20&\underline{13/20}&5/20&5/20&--&--&--&0/20\\
    &$\pi_0$ (AtomBridge)&19/20&\underline{16/20}&11/20&\textbf{14/20}&\textbf{18/20}&--&--&--&\underline{6/20}\\
    &$\pi_{0.5}$ (AtomBridge)&19/20&\textbf{17/20}&\textbf{14/20}&\textbf{14/20}&\underline{9/20}&--&--&--&\textbf{8/20}\\
    \cline{1-11}
    \multirow{6}{*}{E}&$\pi_0$ (vanilla)&17/20&0/20&0/20&0/20&0/20&0/20&0/20&0/20&0/20\\
    &$\pi_{0.5}$ (vanilla)&17/20&0/20&0/20&0/20&0/20&0/20&0/20&0/20&0/20\\
    &$\pi_0$ (MP)&17/20&1/20&1/20&0/20&\textbf{13/20}&1/20&0/20&0/20&0/20\\
    &$\pi_{0.5}$ (MP)&18/20&1/20&\underline{4/20}&2/20&\underline{8/20}&0/20&0/20&3/20&0/20\\
    &$\pi_0$ (AtomBridge)&16/20&\underline{6/20}&3/20&\underline{12/20}&4/20&\underline{15/20}&\underline{8/20}&\underline{13/20}&\underline{2/20}\\
    &$\pi_{0.5}$ (AtomBridge)&18/20&\textbf{13/20}&\textbf{12/20}&\textbf{17/20}&7/20&\textbf{18/20}&\textbf{10/20}&\textbf{16/20}&\textbf{5/20}\\
    \cline{1-11}
    \multirow{6}{*}{F}&$\pi_0$ (vanilla)&19/20&0/20&\underline{15/20}&0/20&0/20&0/20&0/20&0/20&0/20\\
    &$\pi_{0.5}$ (vanilla)&19/20&0/20&13/20&0/20&0/20&0/20&0/20&0/20&0/20\\
    &$\pi_0$ (MP)&19/20&5/20&11/20&\underline{12/20}&1/20&3/20&2/20&0/20&0/20\\
    &$\pi_{0.5}$ (MP)&20/20&2/20&13/20&5/20&9/20&0/20&2/20&0/20&0/20\\
    &$\pi_0$ (AtomBridge)&19/20&\underline{16/20}&10/20&\underline{12/20}&\underline{16/20}&\underline{7/20}&\underline{12/20}&\textbf{10/20}&\underline{2/20}\\
    &$\pi_{0.5}$ (AtomBridge)&20/20&\textbf{20/20}&\textbf{16/20}&\textbf{16/20}&\textbf{20/20}&\textbf{8/20}&\textbf{18/20}&\textbf{10/20}&\textbf{4/20}\\
  \bottomrule
\end{tabular}
\endgroup
\caption{Per-atomic-task and full-sequence success rates for Tasks A--F. Dashes indicate atomic-task positions not present in the five-step tasks A--D.}
\label{tab:scitasks-result}
\end{table*}

\section{Experiments}
\label{sec:exp}
\subsection{Experimental Setup}
\noindent
\textbf{Baseline Settings.}
We use two VLA models, $\pi_0$~\cite{black2024pi_0} and $\pi_{0.5}$~\cite{black2025pi_}, as base policies. 
For each policy, we construct two baselines. 
\textit{Vanilla} executes the atomic-task policies consecutively without inserting a transition. 
\textit{MP} inserts transitions generated from conventional motion planning by bidirectional joint-space RRT-Connect~\cite{jung2025planning} between consecutive atomic tasks, but uses no LLM agent or agent-generated code. 
Its ready state is randomly sampled from the initial states of the matched next atomic task.
If the task description cannot be matched, it will randomly sample a initial state from the memory as the ready state.
\textit{AtomBridge} uses the same base policies, and uses Qwen3.5-397B-A17B as agents' base model.
Its plugin-equipped variants are denoted AtomBridge-$\pi_0$ and AtomBridge-$\pi_{0.5}$.

\noindent
\textbf{Data Collection.}
Across simulation and real-world settings, we collect 100 demonstrations for each of 19 atomic tasks. 
The tasks manipulate scientific instruments and consumables such as centrifuges, thermal cyclers, PCR plates, and centrifuge tubes, etc. The simulation scenes are built on the Mujoco engine by using Autobio~\cite{lan2025autobio} 3D assets. 
We compose these atomic tasks into seven long-horizon task suites (A-G in Fig.~\ref{figs:scitask}). 
Tasks C, D, and G represent experiment-execution workflows. 
Tasks A and B represent experiment-collection workflows. 
Tasks E and F represent complete experimental workflows. 
Detailed collection protocols, state randomization, and task-success annotations are provided in the supplementary material.

\noindent
\textbf{Evaluation Metrics.}
For each method and long-horizon task suite, we run $N=20$ episodes without resetting the environment between atomic tasks. 
Atomic task $i$ in episode $n$ is successful, $s_i^{(n)}=1$, only if its task-specific state predicates are satisfied at both the start and the end of execution. 
Otherwise, $s_i^{(n)}=0$. 
For example, \texttt{open lid} requires a closed, locked lid at the start and a fully open lid at the end.

For a suite of $K$ atomic tasks, the entries under \textit{Success Rate Per Atomic Task} in Tab.~\ref{tab:scitasks-result} are computed as
\begin{equation}
    \mathrm{SR}_i = \frac{1}{N}\sum_{n=1}^{N}s_i^{(n)}.
\end{equation}
Thus, an entry of $a/20$ means that atomic task $i$ succeeded in $a$ of the 20 continuous episodes. 
The \textit{Full-Seq. Success} column requires every atomic task in the same episode to succeed:
\begin{equation}
    \mathrm{FSR} = \frac{1}{N}\sum_{n=1}^{N}
    \prod_{i=1}^{K}s_i^{(n)}.
\end{equation}
Accordingly, $a/20$ in this column denotes $a$ complete task sequences with no failed atomic task.

\subsection{Scientific Operation Results}
Tab.~\ref{tab:scitasks-result} shows that AtomBridge improves most later-step atomic-task success rates relative to the matched transition baseline. 
Across the 60 Atomic 2 onward comparisons, 53 increase, two are unchanged, and five decrease. 
The decreases occur in isolated task-model pairs, including Atomic E.5, where AtomBridge-$\pi_0$ decreases from 13/20 to 4/20. 

The full-sequence results show a clearer long-horizon benefit. 
All baseline configurations obtain 0/20 full-sequence success, whereas AtomBridge obtains non-zero full-sequence success in 11 of the 12 task-model configurations, ranging from 2/20 to 14/20. 
AtomBridge-$\pi_{0.5}$ achieves a higher full-sequence success rate than AtomBridge-$\pi_0$ in five of the six long tasks. 
Task B is the exception, where AtomBridge-$\pi_0$ reaches 14/20, and AtomBridge-$\pi_{0.5}$ reaches 7/20. 
These results indicate that AtomBridge can improve execution continuity across atomic tasks, while also showing that transition handling alone does not eliminate failures in individual atomic policies or guarantee end-to-end success.

We further evaluated physical execution in the thermal-cycler scene of Task G (Fig.~\ref{figs:scitask}). 
As shown in Tab.~\ref{tab:scitasks-result-real}, AtomBridge-$\pi_{0.5}$ succeeded in 4/10 trials for Atomic Tasks 3-5 and for the full sequence, whereas the vanilla policy achieved 0/10 in each of these cases. 
Both methods succeeded in 5/10 trials on Atomic Task 2, consistent with the close match between the end state of Atomic Task 1 and the initial state of Atomic Task 2 in the collected demonstrations. 
These real-environment results support the benefit of AtomBridge specifically at task boundaries that require a state transition.

\begin{table}[!ht]
    \setlength{\tabcolsep}{1.3mm}
    \centering
    \begingroup
    % \scriptsize
    \small
    \begin{tabular}{l|ccccc|c}
    \toprule
    \multirow{2}{*}{Method}&\multicolumn{5}{c|}{Success Rate Per Atomic Task}&\multirow{2}{*}{\makecell{Full-Seq.\\Success}}\\
    \cline{2-6}
    &1&2&3&4&5&\\
    \midrule
    $\pi_{0.5}$ (vanilla)&5/10&5/10&0/10&0/10&0/10&0/10\\
    $\pi_{0.5}$ (AtomBridge) & 5/10 & 5/10 & \textbf{4/10} & \textbf{4/10} & \textbf{4/10} &\textbf{4/10}\\
  \bottomrule
\end{tabular}
\endgroup
\caption{Per-atomic-task and full-sequence success rates for the five-step Task G in a real environment.}
\label{tab:scitasks-result-real}
\end{table}

\subsection{Completion Model Evaluation}
\begin{figure*}[t]
    \centering
    \includegraphics[width=\textwidth]{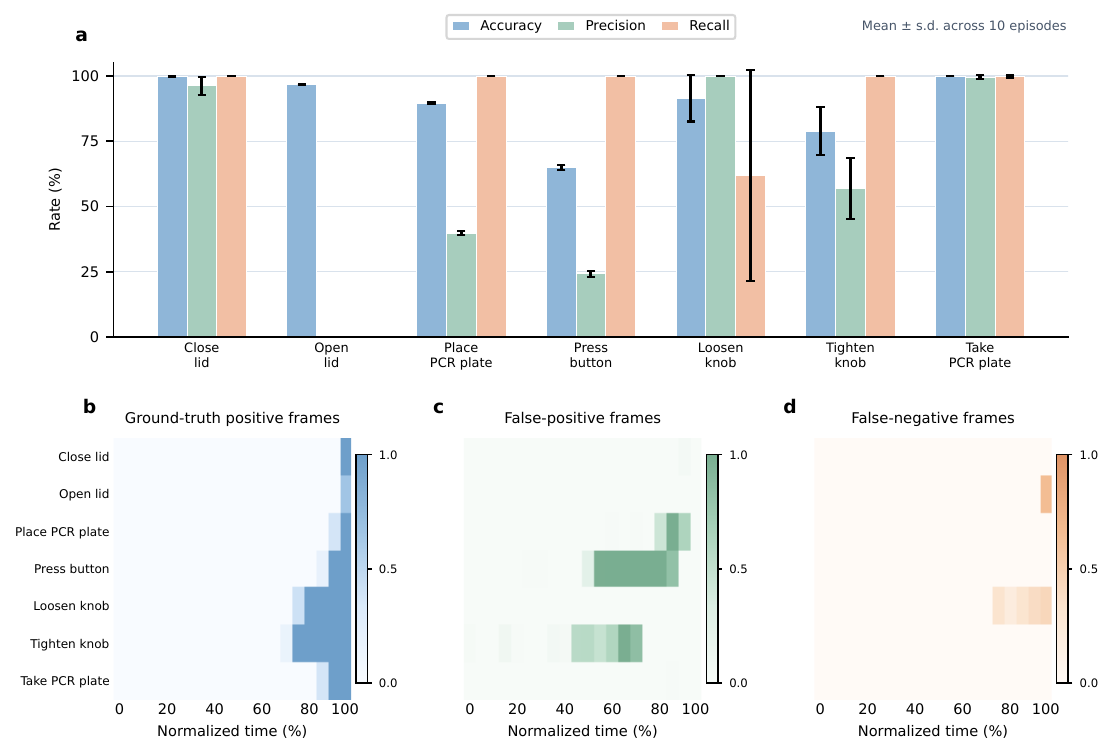}
    \caption{Completion-model evaluation over ten episodes for each thermal-cycler atomic task. \textbf{a}, Mean frame-level accuracy, precision, and recall. Error bars denote standard deviations across episodes. \textbf{b--d}, Fractions of ground-truth positive, false-positive, and false-negative frames in 20 normalized execution-time bins.}
    \label{fig:completion-switch-eval}
\end{figure*}

We evaluate the completion model on 10 episodes for each of the seven thermal-cycler atomic tasks (70 episodes and 34,872 stride-10 frames).  
The model receives the initial frame, current frame, and task description. 
Simulator success predicates provide frame-level labels.  
For all reported experiments, we use the shared default threshold of $0.05$.  
Figure~\ref{fig:completion-switch-eval} reports mean per-episode accuracy, precision, and recall, together with the temporal distributions of ground-truth positive, false-positive, and false-negative frames over 20 normalized execution bins.

Across all frames, the model achieved 92.2\% accuracy, 59.1\% precision, and 84.7\% recall.  
It was nearly exact for persistent end states: closing the lid reached 99.8\% accuracy with 96.2\% precision and 100\% recall, and removing the PCR plate reached 99.9\% accuracy with 99.6\% precision and 99.9\% recall.  
In contrast, placement, button pressing, and knob tightening retained perfect recall but had lower precision, indicating early completion predictions rather than missed completions.  

The open-lid task obtained 96.7\% accuracy but zero precision and recall.  
This is because collection terminates immediately after the lid opens, leaving only a short positive segment and no post-completion robot motion. 

The heatmaps localize these errors.  
Ground-truth positives concentrate at the end of every trajectory.  
False positives occur before this terminal region for PCR-plate placement, button pressing, and knob tightening, whereas false negatives concentrate at the end of lid opening and knob loosening.  
Thus, the completion signal is reliable for detecting a candidate task boundary, but ambiguous visual changes require temporal confirmation.  
All reported experiments use the default threshold together with a short consecutive-positive window.  
Increasing this window can mitigate premature predictions and we set it to 10 for all experiments.  
In principle, selecting task-type-specific window lengths could further improve intervention effectiveness.
If the completion model does not fire, as in the open-lid case, the task-specific maximum execution time triggers the intervention switch as a fallback.

\begin{table}[!t]
\centering
\small
\begin{tabular}{l|cc|cc}
\toprule
\multirow{2}{*}{Method} & \multicolumn{2}{c|}{Key Tools} & \multicolumn{2}{c}{Eval.} \\
\cline{2-5}
& \makecell{Ready\\retrieval} & \makecell{Collision\\Check} & \makecell{D.2\\Success} & \makecell{Contact\\Pair-Samples} \\
\midrule
No-check & $\checkmark$ & $\times$ & 10/20 & 165.35 \\
No-retrieval & $\times$ & $\checkmark$ & 0/20 & 82.50 \\
check+retrieval & $\checkmark$ & $\checkmark$ & \textbf{19/20} & 18.00 \\
\bottomrule
\end{tabular}
\caption{Ablation results on the long task D.1$\rightarrow$D.2. A check mark indicates an enabled tool.}
\label{tab:ablation_long_task}
\end{table}

\begin{table}[t]
\small
\setlength{\tabcolsep}{1.8mm}
\begin{tabular}{lccc}
\toprule
Model & \makecell{Transition\\Generation\\Time} & \makecell{Full-Seq.\\Time} & \makecell{D.2\\Success} \\
\midrule
GPT-5.2 & 38s (78\%) & 49s (100\%) & 10/10 \\
Qwen3.5-397B-A17B & 185s (81\%) & 227s (100\%) & 8/10 \\
Qwen3.5-35B-A3B & 69s (60\%) & 115s (100\%) & 7/10 \\
Qwen3.5-9B & 55s (56\%) & 99s (100\%) & 5/10 \\
\bottomrule
\end{tabular}
\caption{Comparison of the proprietary GPT-5.2 model and three open-source Qwen3.5 variants on the D.1$\rightarrow$D.2 task.  Parenthetical values report each duration as a proportion of the complete long-horizon task time, and D.2 success rates are measured over 10 trials.}
\label{tab:llm_inference_time}
\end{table}

\subsection{Ablation Studies}
We study the D.1$\rightarrow$D.2 transition because it combines instrument opening with high-precision placement.
As shown in Tab.~\ref{tab:ablation_long_task}, the ablation compares the complete pipeline with No-check and No-retrieval variants.
Because collision check and ready-state retrieval are key tools in the whole pipeline.
For each setting, we collect 20 valid trials in which D.1 succeeds and report the conditional D.2 success rate and transition \textit{Contact Pair-Samples}.
The conditional D.2 success rate is the success rate of D.2 under the condition that D.1 succeeds.
Meanwhile, \textit{Contact Pair-Samples} is a metric to quantify collision severity during transitions. 
This metric counts the sampled contact pairs between the robot arm and environment objects across execution time. 
It aggregates contact incidence and duration because multiple pairs may be recorded at each sampled time step. 
Lower values therefore indicate fewer accumulated contact pair samples under this sampling scheme, not a smaller physical collision area or guaranteed safety.
This compound comparison measures ready-state target guidance as a whole and does not isolate retrieval alone.
The complete pipeline achieves the highest conditional D.2 success rate and the fewest contact pair samples.

\subsection{Latency and Generation Quality}
As shown in Tab.~\ref{tab:llm_inference_time}, we compare the proprietary GPT-5.2 model with three open-source Qwen3.5 variants on D.1$\rightarrow$D.2 over 10 trials. 
VLA inference uses a single NVIDIA 5070 GPU. 
All LLMs are accessed through remote inference services, so the reported transition generation times include both server-side inference and network communication latency. 
GPT-5.2 achieves the highest observed D.2 success rate (10/10) and the shortest transition generation time (38s), which accounts for 78\% of the 49s task time. 
Within the open-source Qwen3.5 family, D.2 success decreases from 8/10 for the 397B model to 7/10 for the 35B model and 5/10 for the 9B model. 
Transition generation time also decreases from 185s to 69s and 55s, respectively. 
These open-source results indicate a latency-quality trade-off within the Qwen3.5 family. 
Because each model is evaluated over only 10 trials, the cross-family results are preliminary. 
Moreover, the reported timings represent end-to-end remote deployment measurements rather than hardware-normalized model-inference times.

\section{Conclusion}
This paper proposes AtomBridge, an inference plugin that composes fine-tuned atomic VLA policies without additional VLA fine-tuning or long-horizon demonstrations. 
Across the evaluated simulated sequences, AtomBridge improves per-step success for most subsequent atomic tasks and increases execution continuity. 
Ultimately improving the execution success rate for full-sequence tasks.

However, AtomBridge does not explicitly reason about or restore non-robot states, such as the states of manipulated objects and laboratory instruments.
Future work could extend AtomBridge to perceive and restore object and instrument states, thereby addressing a broader portion of the skill-chaining gap. 
In addition, it is necessary to reduce remote LLM inference latency and improve the robustness of visual completion detection under occlusion and real-world domain shifts.

\bibliography{refs}

@inproceedings{radford2021learning,
  title={Learning transferable visual models from natural language supervision},
  author={Radford, Alec and Kim, Jong Wook and Hallacy, Chris and Ramesh, Aditya and Goh, Gabriel and Agarwal, Sandhini and Sastry, Girish and Askell, Amanda and Mishkin, Pamela and Clark, Jack and others},
  booktitle={International conference on machine learning},
  pages={8748--8763},
  year={2021},
  organization={PMLR}
}

@inproceedings{gupta2020relay,
  title={Relay Policy Learning: Solving Long-Horizon Tasks via Imitation and Reinforcement Learning},
  author={Gupta, Abhishek and Kumar, Vikash and Lynch, Corey and Levine, Sergey and Hausman, Karol},
  booktitle={Conference on Robot Learning},
  pages={1025--1037},
  year={2020},
  organization={PMLR}
}

@article{zhang2024vision,
  title={Vision-language models for vision tasks: A survey},
  author={Zhang, Jingyi and Huang, Jiaxing and Jin, Sheng and Lu, Shijian},
  journal={IEEE transactions on pattern analysis and machine intelligence},
  volume={46},
  number={8},
  pages={5625--5644},
  year={2024},
  publisher={IEEE}
}

@article{wang2023scientific,
  title={Scientific discovery in the age of artificial intelligence},
  author={Wang, Hanchen and Fu, Tianfan and Du, Yuanqi and Gao, Wenhao and Huang, Kexin and Liu, Ziming and Chandak, Payal and Liu, Shengchao and Van Katwyk, Peter and Deac, Andreea and others},
  journal={Nature},
  volume={620},
  number={7972},
  pages={47--60},
  year={2023},
  publisher={Nature Publishing Group UK London}
}

@article{koscher2023autonomous,
  title={Autonomous, multiproperty-driven molecular discovery: From predictions to measurements and back},
  author={Koscher, Brent A and Canty, Richard B and McDonald, Matthew A and Greenman, Kevin P and McGill, Charles J and Bilodeau, Camille L and Jin, Wengong and Wu, Haoyang and Vermeire, Florence H and Jin, Brooke and others},
  journal={Science},
  volume={382},
  number={6677},
  pages={eadi1407},
  year={2023},
  publisher={American Association for the Advancement of Science}
}

@article{szymanski2023autonomous,
  title={An autonomous laboratory for the accelerated synthesis of novel materials},
  author={Szymanski, Nathan J and Rendy, Bernardus and Fei, Yuxing and Kumar, Rishi E and He, Tanjin and Milsted, David and McDermott, Matthew J and Gallant, Max and Cubuk, Ekin Dogus and Merchant, Amil and others},
  journal={Nature},
  volume={624},
  number={7990},
  pages={86--91},
  year={2023},
  publisher={Nature Publishing Group UK London}
}

@article{boiko2023autonomous,
  title={Autonomous chemical research with large language models},
  author={Boiko, Daniil A and MacKnight, Robert and Kline, Ben and Gomes, Gabe},
  journal={Nature},
  volume={624},
  number={7992},
  pages={570--578},
  year={2023},
  publisher={Nature Publishing Group UK London}
}

@article{jumper2021highly,
  title={Highly accurate protein structure prediction with AlphaFold},
  author={Jumper, John and Evans, Richard and Pritzel, Alexander and Green, Tim and Figurnov, Michael and Ronneberger, Olaf and Tunyasuvunakool, Kathryn and Bates, Russ and {\v{Z}}{\'\i}dek, Augustin and Potapenko, Anna and others},
  journal={Nature},
  volume={596},
  number={7873},
  pages={583--589},
  year={2021},
  publisher={Nature Publishing Group UK London}
}

@article{bi2023accurate,
  title={Accurate medium-range global weather forecasting with 3D neural networks},
  author={Bi, Kaifeng and Xie, Lingxi and Zhang, Hengheng and Chen, Xin and Gu, Xiaotao and Tian, Qi},
  journal={Nature},
  volume={619},
  number={7970},
  pages={533--538},
  year={2023},
  publisher={Nature Publishing Group UK London}
}

@article{coley2019robotic,
  title={A robotic platform for flow synthesis of organic compounds informed by AI planning},
  author={Coley, Connor W and Thomas III, Dale A and Lummiss, Justin AM and Jaworski, Jonathan N and Breen, Christopher P and Schultz, Victor and Hart, Travis and Fishman, Joshua S and Rogers, Luke and Gao, Hanyu and others},
  journal={Science},
  volume={365},
  number={6453},
  pages={eaax1566},
  year={2019},
  publisher={American Association for the Advancement of Science}
}

@inproceedings{kim2025openvla,
  title={OpenVLA: An Open-Source Vision-Language-Action Model},
  author={Kim, Moo Jin and Pertsch, Karl and Karamcheti, Siddharth and Xiao, Ted and Balakrishna, Ashwin and Nair, Suraj and Rafailov, Rafael and Foster, Ethan P and Sanketi, Pannag R and Vuong, Quan and others},
  booktitle={Conference on Robot Learning},
  pages={2679--2713},
  year={2025},
  organization={PMLR}
}

@article{beyer2024paligemma,
  title={PaliGemma: A versatile 3B VLM for transfer},
  author={Beyer, Lucas and Steiner, Andreas and Pinto, Andr{\'e} Susano and Kolesnikov, Alexander and Wang, Xiao and Salz, Daniel and Neumann, Maxim and Alabdulmohsin, Ibrahim and Tschannen, Michael and Bugliarello, Emanuele and others},
  journal={CoRR},
  year={2024}
}

@article{lan2025autobio,
  title={Autobio: A simulation and benchmark for robotic automation in digital biology laboratory},
  author={Lan, Zhiqian and Jiang, Yuxuan and Wang, Ruiqi and Xie, Xuanbing and Zhang, Rongkui and Zhu, Yicheng and Li, Peihang and Yang, Tianshuo and Chen, Tianxing and Gao, Haoyu and others},
  journal={arXiv preprint arXiv:2505.14030},
  year={2025}
}

@article{black2024pi_0,
  title={$\pi_ {0} $: A Vision-Language-Action Flow Model for General Robot Control},
  author={Black, Kevin and Brown, Noah and Driess, Danny and Esmail, Adnan and Equi, Michael and Finn, Chelsea and Fusai, Niccolo and Groom, Lachy and Hausman, Karol and Ichter, Brian and others},
  journal={arXiv preprint arXiv:2410.24164},
  year={2024}
}

@article{macleod2020self,
  title={Self-driving laboratory for accelerated discovery of thin-film materials},
  author={MacLeod, Benjamin P and Parlane, Fraser GL and Morrissey, Thomas D and H{\"a}se, Florian and Roch, Lo{\"\i}c M and Dettelbach, Kevan E and Moreira, Raphaell and Yunker, Lars PE and Rooney, Michael B and Deeth, Joseph R and others},
  journal={Science Advances},
  volume={6},
  number={20},
  pages={eaaz8867},
  year={2020},
  publisher={American Association for the Advancement of Science}
}

@article{zhao2021discovery,
  title={Discovery of temperature-induced stability reversal in perovskites using high-throughput robotic learning},
  author={Zhao, Yicheng and Zhang, Jiyun and Xu, Zhengwei and Sun, Shijing and Langner, Stefan and Hartono, Noor Titan Putri and Heumueller, Thomas and Hou, Yi and Elia, Jack and Li, Ning and others},
  journal={Nature Communications},
  volume={12},
  number={1},
  pages={2191},
  year={2021},
  publisher={Nature Publishing Group UK London}
}

@article{burger2020mobile,
  title={A mobile robotic chemist},
  author={Burger, Benjamin and Maffettone, Phillip M and Gusev, Vladimir V and Aitchison, Catherine M and Bai, Yang and Wang, Xiaoyan and Li, Xiaobo and Alston, Ben M and Li, Buyi and Clowes, Rob and others},
  journal={Nature},
  volume={583},
  number={7815},
  pages={237--241},
  year={2020},
  publisher={Nature Publishing Group UK London}
}

@article{king2009automation,
  title={The automation of science},
  author={King, Ross D and Rowland, Jem and Oliver, Stephen G and Young, Michael and Aubrey, Wayne and Byrne, Emma and Liakata, Maria and Markham, Magdalena and Pir, Pinar and Soldatova, Larisa N and others},
  journal={Science},
  volume={324},
  number={5923},
  pages={85--89},
  year={2009},
  publisher={American Association for the Advancement of Science}
}

@article{chi2025diffusion,
  title={Diffusion policy: Visuomotor policy learning via action diffusion},
  author={Chi, Cheng and Xu, Zhenjia and Feng, Siyuan and Cousineau, Eric and Du, Yilun and Burchfiel, Benjamin and Tedrake, Russ and Song, Shuran},
  journal={The International Journal of Robotics Research},
  volume={44},
  number={10-11},
  pages={1684--1704},
  year={2025},
  publisher={Sage Publications Sage UK: London, England}
}

@inproceedings{liurdt,
  title={RDT-1B: a Diffusion Foundation Model for Bimanual Manipulation},
  author={Liu, Songming and Wu, Lingxuan and Li, Bangguo and Tan, Hengkai and Chen, Huayu and Wang, Zhengyi and Xu, Ke and Su, Hang and Zhu, Jun},
  booktitle={The Thirteenth International Conference on Learning Representations},
  year={2025}
}

@article{volk2023alphaflow,
  title={AlphaFlow: autonomous discovery and optimization of multi-step chemistry using a self-driven fluidic lab guided by reinforcement learning},
  author={Volk, Amanda A and Epps, Robert W and Yonemoto, Daniel T and Masters, Benjamin S and Castellano, Felix N and Reyes, Kristofer G and Abolhasani, Milad},
  journal={Nature Communications},
  volume={14},
  number={1},
  pages={1403},
  year={2023},
  publisher={Nature Publishing Group UK London}
}

@article{dietterich2000hierarchical,
  title={Hierarchical reinforcement learning with the MAXQ value function decomposition},
  author={Dietterich, Thomas G},
  journal={Journal of Artificial Intelligence Research},
  volume={13},
  pages={227--303},
  year={2000}
}

@inproceedings{chen2023sequential,
  title={Sequential Dexterity: Chaining Dexterous Policies for Long-Horizon Manipulation},
  author={Chen, Yuanpei and Wang, Chen and Fei-Fei, Li and Liu, Karen},
  booktitle={Conference on Robot Learning},
  pages={3809--3829},
  year={2023},
  organization={PMLR}
}

@article{konidaris2009skill,
  title={Skill discovery in continuous reinforcement learning domains using skill chaining},
  author={Konidaris, George and Barto, Andrew},
  journal={Advances in neural information processing systems},
  volume={22},
  year={2009}
}

@article{clegg2018learning,
  title={Learning to dress: Synthesizing human dressing motion via deep reinforcement learning},
  author={Clegg, Alexander and Yu, Wenhao and Tan, Jie and Liu, C Karen and Turk, Greg},
  journal={ACM Transactions on Graphics (TOG)},
  volume={37},
  number={6},
  pages={1--10},
  year={2018},
  publisher={ACM New York, NY, USA}
}

@article{dai2025adaptive,
  title={Adaptive AI decision interface for autonomous electronic material discovery},
  author={Dai, Yahao and Chan, Henry and Vriza, Aikaterini and Fan, Jingyuan and Kim, Fredrick and Wang, Yunfei and Liu, Wei and Shan, Naisong and Xu, Jing and Weires, Max and others},
  journal={Nature Chemical Engineering},
  pages={1--11},
  year={2025},
  publisher={Nature Publishing Group US New York}
}

@article{coley2020autonomous,
  title={Autonomous discovery in the chemical sciences part I: Progress},
  author={Coley, Connor W and Eyke, Natalie S and Jensen, Klavs F},
  journal={Angewandte Chemie International Edition},
  volume={59},
  number={51},
  pages={22858--22893},
  year={2020},
  publisher={Wiley Online Library}
}

@article{ochiai2025automating,
  title={Automating Care by Self-maintainability for Full Laboratory Automation},
  author={Ochiai, Koji and Tahara-Arai, Yuya and Kato, Akari and Kaizu, Kazunari and Kariyazaki, Hirokazu and Umeno, Makoto and Takahashi, Koichi and Kanda, Genki N and Ozaki, Haruka},
  journal={arXiv preprint arXiv:2501.05789},
  year={2025}
}

@inproceedings{black2025pi_,
  title={$\pi_ {0.5} $: a Vision-Language-Action Model with Open-World Generalization},
  author={Black, Kevin and Brown, Noah and Darpinian, James and Dhabalia, Karan and Driess, Danny and Esmail, Adnan and Equi, Michael Robert and Finn, Chelsea and Fusai, Niccolo and Galliker, Manuel Y and others},
  booktitle={9th Annual Conference on Robot Learning},
  year={2025}
}

@inproceedings{fan2025long,
  title={Long-VLA: Unleashing Long-Horizon Capability of Vision Language Action Model for Robot Manipulation},
  author={Fan, Yiguo and Bai, Shuanghao and Tong, Xinyang and Ding, Pengxiang and Zhu, Yuyang and Lu, Hongchao and Dai, Fengqi and Zhao, Wei and Liu, Yang and Huang, Siteng and others},
  booktitle={Conference on Robot Learning},
  pages={2018--2037},
  year={2025},
  organization={PMLR}
}

@article{yang2025lohovla,
  title={LoHoVLA: A Unified Vision-Language-Action Model for Long-Horizon Embodied Tasks},
  author={Yang, Yi and Sun, Jiaxuan and Kou, Siqi and Wang, Yihan and Deng, Zhijie},
  journal={arXiv preprint arXiv:2506.00411},
  year={2025}
}

@article{macarron2011impact,
  title={Impact of high-throughput screening in biomedical research},
  author={Macarron, Ricardo and Banks, Martyn N and Bojanic, Dejan and Burns, David J and Cirovic, Dragan A and Garyantes, Tina and Green, Darren VS and Hertzberg, Robert P and Janzen, William P and Paslay, Jeff W and others},
  journal={Nature Reviews Drug Discovery},
  volume={10},
  number={3},
  pages={188--195},
  year={2011},
  publisher={Nature Publishing Group UK London}
}

@article{boyd2002robotic,
  title={Robotic laboratory automation},
  author={Boyd, James},
  journal={Science},
  volume={295},
  number={5554},
  pages={517--518},
  year={2002},
  publisher={American Association for the Advancement of Science}
}

@article{novikov2025alphaevolve,
  title={AlphaEvolve: A coding agent for scientific and algorithmic discovery},
  author={Novikov, Alexander and V{\~u}, Ng{\^a}n and Eisenberger, Marvin and Dupont, Emilien and Huang, Po-Sen and Wagner, Adam Zsolt and Shirobokov, Sergey and Kozlovskii, Borislav and Ruiz, Francisco JR and Mehrabian, Abbas and others},
  journal={arXiv preprint arXiv:2506.13131},
  year={2025}
}

@inproceedings{mishra2023generative,
  title={Generative skill chaining: Long-horizon skill planning with diffusion models},
  author={Mishra, Utkarsh Aashu and Xue, Shangjie and Chen, Yongxin and Xu, Danfei},
  booktitle={Conference on Robot Learning},
  pages={2905--2925},
  year={2023},
  organization={PMLR}
}

@article{jung2025planning,
  title={From planning to policy: distilling \texttt{Skill-RRT} for long-horizon prehensile and non-prehensile manipulation},
  author={Jung, Haewon and Lee, Donguk and Park, Haecheol and Kim, JunHyeop and Kim, Beomjoon},
  journal={arXiv e-prints},
  pages={arXiv--2502},
  year={2025}
}

@inproceedings{lee2022adversarial,
  title={Adversarial Skill Chaining for Long-Horizon Robot Manipulation via Terminal State Regularization},
  author={Lee, Youngwoon and Lim, Joseph J and Anandkumar, Anima and Zhu, Yuke},
  booktitle={Conference on Robot Learning},
  pages={406--416},
  year={2022},
  organization={PMLR}
}

@InProceedings{pmlr-v87-ebert18a,
  title = 	 {Robustness via Retrying: Closed-Loop Robotic Manipulation with Self-Supervised Learning},
  author =       {Ebert, Frederik and Dasari, Sudeep and Lee, Alex X. and Levine, Sergey and Finn, Chelsea},
  booktitle = 	 {Proceedings of The 2nd Conference on Robot Learning},
  pages = 	 {983--993},
  year = 	 {2018},
  editor = 	 {Billard, Aude and Dragan, Anca and Peters, Jan and Morimoto, Jun},
  volume = 	 {87},
  series = 	 {Proceedings of Machine Learning Research},
  month = 	 {29--31 Oct},
  publisher =    {PMLR},
  url = 	 {https://proceedings.mlr.press/v87/ebert18a.html}
}

@ARTICLE{10328058,
  author={Yokoyama, Naoki and Clegg, Alex and Truong, Joanne and Undersander, Eric and Yang, Tsung-Yen and Arnaud, Sergio and Ha, Sehoon and Batra, Dhruv and Rai, Akshara},
  journal={IEEE Robotics and Automation Letters}, 
  title={ASC: Adaptive Skill Coordination for Robotic Mobile Manipulation}, 
  year={2024},
  volume={9},
  number={1},
  pages={779-786},
  doi={10.1109/LRA.2023.3336109}}

@InProceedings{pmlr-v205-sung23a,
  title = 	 {Learning to Correct Mistakes: Backjumping in Long-Horizon Task and Motion Planning},
  author =       {Sung, Yoonchang and Wang, Zizhao and Stone, Peter},
  booktitle = 	 {Proceedings of The 6th Conference on Robot Learning},
  pages = 	 {2115--2124},
  year = 	 {2023},
  editor = 	 {Liu, Karen and Kulic, Dana and Ichnowski, Jeff},
  volume = 	 {205},
  series = 	 {Proceedings of Machine Learning Research},
  month = 	 {14--18 Dec},
  publisher =    {PMLR},
  url = 	 {https://proceedings.mlr.press/v205/sung23a.html}
}

@ARTICLE{10400863,
  author={Hou, Yiwen and Ma, Jinming and Sun, Haoyuan and Wu, Feng},
  journal={IEEE Robotics and Automation Letters}, 
  title={Effective Offline Robot Learning With Structured Task Graph}, 
  year={2024},
  volume={9},
  number={4},
  pages={3633-3640},
  doi={10.1109/LRA.2024.3354620}}

@article{ren2026labvla,
  title={LabVLA: Grounding Vision-Language-Action Models in Scientific Laboratories},
  author={Ren, Baochang and Liu, Xinjie and Chen, Xi and Liu, Yanshuo and Li, Chenxi and Gao, Daqi and Su, Zeqin and Xing, Jintao and Xue, Zirui and Li, Rui and others},
  journal={arXiv preprint arXiv:2606.13578},
  year={2026}
}

\end{document}